\documentclass[letterpaper, 10 pt, journal, twoside]{ieeetran}
\usepackage{multirow}
\usepackage{cite}
\usepackage{amsmath,amssymb,amsfonts}
\usepackage{algorithmic}
\usepackage{graphicx}
\usepackage{textcomp}
\usepackage{xcolor}
\usepackage{comment}
\usepackage{makecell} 
\usepackage{siunitx}
\usepackage{subcaption}
\usepackage{dblfloatfix}

\title{A Virtual Mechanical Interaction Layer
Enables Resilient Human-to-Robot Object Handovers
}

\author{Omar Faris$^{1}$, Sławomir K. Tadeja$^{2}$, and Fulvio Forni$^{1}$

\thanks{Manuscript received: November, 10, 2025; Revised January, 22, 2026; Accepted April, 4, 2026.}
\thanks{This paper was recommended for publication by Editor Angelika Peer upon evaluation of the Associate Editor and Reviewers’ comments.
This work was supported by the Engineering and Physical Sciences Research Council and AgriFoRwArdS CDT [EP/S023917/1].}
\thanks{$^{1}$Omar Faris and Fulvio Forni are with the Department of Engineering, University of Cambridge, UK.}
\thanks{$^{2}$Sławomir K. Tadeja is with the Department of Mechanical Engineering, Massachusetts Institute of Technology, USA}
\thanks{Digital Object Identifier (DOI): see top of this page.}
}

\begin{document}

\maketitle

\begin{abstract}

Object handover is a common form of interaction widely used in collaborative tasks. However, achieving it efficiently remains a challenge. We address the problem of ensuring resilient robotic actions that can adapt to complex changes in object pose during human-to-robot object handovers. We propose the use of Virtual Model Control to create an interaction layer that controls the robot and adapts to the dynamic changes in the handover process. Additionally, we propose using augmented reality to facilitate bidirectional communication between humans and robots during handovers. Our controller demonstrated high resilience in experimental tests against various sources of uncertainties with a nearly perfect success rate and an average mean approach time of 4 seconds in one-time rotation or translation tests of four different objects, as well as a success rate of $100\%$ in tests with random object motion and robot starting pose. We also performed a user study with 16 participants to understand human preferences for different robot control profiles and augmented reality visuals in object handovers. Our results showed a general preference for the proposed approach with a 98\% success rate and a mean approach time of 4.9 seconds and revealed insights for improving the interaction with the user.

\end{abstract}

\section{Introduction}

\IEEEPARstart{H}{uman}-to-robot object handover is a fundamental task in collaborative manipulation. While humans interact seamlessly and efficiently during object handovers, replicating that in the presence of robots remains difficult \cite{duan2024human, ortenzi2021object}. Researchers have addressed several challenges to enable safe and efficient human-to-robot object handovers, such as recognizing and communicating human and robot intentions \cite{liu2021object, wang2021predicting}, detecting and grasping objects with diverse physical properties \cite{rosenberger2020object, huang2023fed, pang2025stereo}, and ensuring safe contact with the human \cite{castellani2024soft}. A key challenge is enabling robots to adapt and respond to the continuous changes during the handover task, particularly the variability and unpredictability of human motion \cite{ortenzi2021object}. We summarize this capability with the concept of \emph{resilient} handover, in contrast to non-resilient cases that force the human to maintain a static giving pose or adhere to a fixed handover location, thereby imposing unnatural constraints and uncomfortable strain.

Prior research has mostly attempted to achieve resilient human-to-robot handovers by improving grasp generation methods and re-plan trajectories toward the updated target grasp \cite{zhang2023flexible, yang2021reactive, duan2024reactive, yang2022model}.
However, these approaches often treat resilience as a path planning problem, which can be computationally expensive, and ignore the interactive nature of handovers altogether. In contrast, we examine this problem from the perspective of \emph{motor coordination} between the robot and the human/object. We consider the human and robot as a coordinated ensemble regulated by an interaction layer. Our hypothesis is that this layer will shape the feedback loop between the human and the robot, enabling more resilient handover without compromising performance. The goal is to design this algorithmic layer to ensure smooth, responsive, and reactive coordination between the two agents.

\begin{figure}[t!]
  \centering
  \includegraphics[width=0.83\linewidth]{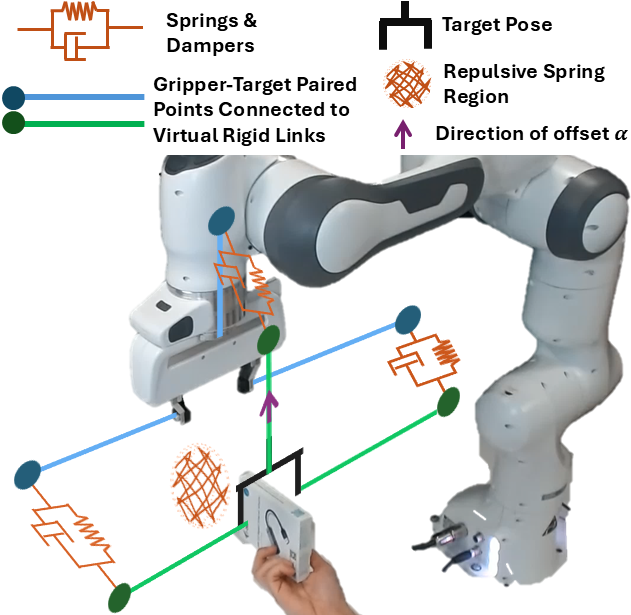}
  \caption{Our interaction layer based on Virtual Model Control. The Main Controller Module consists of Gripper-Target Paired Points (target points are translated by an offset $\alpha$) connected to the gripper and the object/hand via Virtual Rigid Links, with Springs and Dampers between each pair. Repulsive Spring Regions form the Auxiliary Controller Module, with one region here placed in front of the object/hand.}
  \label{fig:main}
\end{figure}

In this work, we develop an interaction layer that takes the form of a ``virtual'' mechanism, as shown in Fig. \ref{fig:main}, following the philosophy of Virtual Model Control (VMC)\cite{pratt2001virtual,larby2024optimal}. VMC has demonstrated promising results in diverse tasks ranging from locomotion \cite{pratt2001virtual,winkler2014path, chen2020virtual}, to pick and place and obstacle avoidance \cite{zhang2024virtual}, and robotic surgery \cite{larby2025collaborative,larby2024optimal}. Rather than re-planning trajectories, we couple the robot and human via virtual mechanical components (links, joints, springs, and dampers) to generate virtual forces that drive the robot's motion and shape its desired reactions. The coordination and motion of the robot are the direct result of this ``mechanical'' interaction, mediated by the virtual components, which enable reactive behavior to changes in the object pose.

Our approach is implemented through Augmented Reality (AR), which provides native hand-tracking capabilities.
 A head-mounted display tracks the user's hand and shares that information with the robot while also visualizing the robot's intended grasping pose, which is continuously adapted to reflect real-time adjustments in the robot's motion towards the object. AR has shown promising results for bidirectional human–robot communication \cite{suzuki2022augmented, 10802282}. In particular, the visualization of the robot's intended grasping pose has previously been shown to improve handover performance \cite{newbury2022visualizing}.

To evaluate our approach, we completed a series of experiments that assessed the resilience of the virtual model control interaction layer against various perturbations during human-to-robot object handovers. We have also conducted a human user study to validate the usability of our solution with different users and investigate the effectiveness of utilizing augmented reality as a bidirectional communication medium. 

\section{Related Works}

In human-to-robot object handovers, the robot and the human continuously influence each other, causing each handover to be different from the other. Therefore, several works have focused on enabling resilient robot actions that react and adapt to the changes during the handovers. Yang et al. \cite{yang2021reactive} presented reactive robot actions by continuously generating updated grasps as the object moved, which was later improved in \cite{yang2022model} by incorporating model predictive control to achieve smoother motion under task constraints. The same principle of continuously updating the grasp pose and re-planning robot actions towards it was also used in \cite{duan2024reactive}. In contrast, Zhang et al. \cite{zhang2023flexible} focused on predicting future grasps to improve resilience during dynamic handovers.

While these approaches show promising results, they achieve resilient robot behaviors by continuously solving a path planning problem towards updated candidate grasps, which often requires computationally expensive models. In contrast, Wang et al. \cite{wang2024genh2r} proposed learning reactive end-to-end visuomotor handover policies from simulations. However, the approach relied on synthetic data, and its transfer to real-world scenarios involved limited experimental validation. In \cite{costanzo2021handover}, visual servoing was utilized to continuously track and react to object features before the grasping action, which required texture-rich objects. The handover scenario was also formulated as a task-space quadratic-programming control problem to enable real-time resilience to changes in object pose \cite{djeha2022human}. In this case, reliable access to the full object state is required. Therefore, there remains a gap in achieving resilient human-to-robot object handovers that naturally accommodate the continuous nature of human-robot interactions. 

\section{Resilient Human-Robot Interaction Layer}

\subsection{Proposed Virtual Model Controller}

We have identified two major factors for enabling resilient human-robot interactions:
\begin{itemize}
\item
While reaching objects within its workspace,
the robot should seamlessly and continuously adapt to position and orientation changes of the object, as long as this is in the hands of the human. 
\item In approaching the object, the robot should be able to negotiate a collision-free path, avoiding both the human and the object. 
\end{itemize}
We construct a virtual model controller consisting of two modules to realize these features. The Main Controller Module generates virtual attraction forces that drive the robot towards the object, enabling responsive robot actions when the object is moved. In contrast, the Auxiliary Controller Module generates region-specific repulsive forces to guarantee collision avoidance. 
These modules do not require the solution of inverse kinematics or the completion of any motion planning computation. The robot's intelligent and reactive behavior is determined by the set of virtual springs, dampers, and linkages defined within the virtual model control framework, effectively reducing complex control to fundamental virtual mechanical interactions.

\subsubsection{Main Controller Module} 

For a two-fingered gripper, this module is represented by three virtual massless pairs of points. Each pair consists of a \emph{gripper} point rigidly connected to the robot's gripper (blue points in Fig. \ref{fig:main}b), a \emph{target} point connected to the object or the hand (green points in Fig. \ref{fig:main}b), and virtual attractive springs and dampers between them. By using multiple connection points we can control the pose of a two-fingered gripper while keeping its fingers at the appropriate grasping target.

In this paper, the target points are rigidly connected to the object. Any changes in the object pose extend or shrink the lengths of virtual springs and dampers, forcing the gripper to move accordingly, typically towards the object. 
The geometry of these  connection points is regulated by the virtual rigid links represented in Fig. \ref{fig:main}b. These extend from gripper and object, with the virtual connection points placed at the end of them.  
Increasing the length of these links results in higher forces when the object is rotated, enabling the gripper to swiftly respond to the object motion with the right orientation. 

Saturated spring force profiles of the form  
\begin{equation}
    \label{eq:force_profile}
    \textbf{F}_{s} = F_{s_{max}} \tanh\!\left(\frac{k_{s}|\textbf{p}|}{F_{s_{max}}}\right)\frac{\textbf{p}}{|\textbf{p}|}
\end{equation}
are used to generate forces between gripper-target paired points. This  prevents the robot from applying excessive forces for large spring extension.
In \eqref{eq:force_profile}, $F_{s_{max}}>0$ is the maximum force that a spring can apply, $k_s>0$ is the spring's stiffness, and $\textbf{p} \in \mathbb{R}^3$ is the displacement vector between paired gripper-target points.

\begin{figure}[t!]
  \vspace{2.2mm} 
  \centering
  \includegraphics[width=0.99\linewidth]{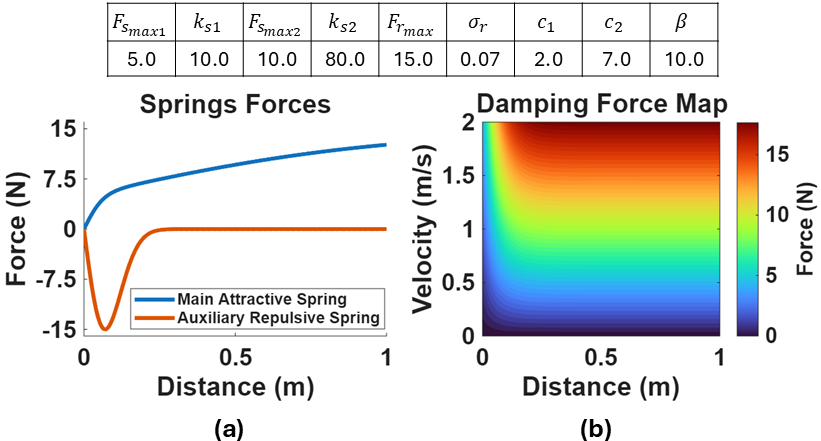}
  \caption{Tuned parameters of the different virtual components used in our virtual model control along with their force profiles: (a) main and auxiliary springs and (b) main dampers. Units are \SI{}{\newton} for force values, \SI{}{\newton\per\meter} for stiffness coefficients, \SI{}{\newton\second\per\meter} for damping coefficients, and \SI{}{\meter} for $\sigma_r$.}
  \label{fig:vmcparams}
\end{figure}

The elastic force generated at each pair of gripper-target points is regulated by two saturated springs.
The first has a large maximal force $F_{s_{max_1}}$ but low stiffness $k_{s_1}$ while the second  relies on a low maximal force $F_{s_{max_2}}$ but high stiffness $k_{s_2}$. The first spring governs large-distance interactions but has limited effects locally. The second spring has limited long-distance interaction but  provides a snap-like mechanism at close range. This is illustrated in Fig. \ref{fig:vmcparams}a.

Virtual dampers generate forces 
of the form
\begin{equation}
\textbf{F}_d = c(|\textbf{p}|)\dot{\textbf{p}}, \ \ \ c(|\textbf{p}|) = c_1 + c_2\tanh(\beta|\textbf{p}|)
\end{equation}
characterized by the position dependent damping $c(|\textbf{p}|) > 0$ for all $s\geq 0$. As before, $\textbf{p}$ refers to the displacement vector between a generic pair of 
gripper-target points. $c_1 \geq 0$ is a constant representing the minimum damping coefficient, $c_1+c_2 > 0 $ is the maximal damping coefficient, and $\beta$ shapes the sensitivity of the damping coefficient to variations in position.  Variable damping offers greater tuning flexibility and has proven effective in experiments.

An offset $\alpha$ ($\alpha = 10$ cm, determined experimentally) is introduced to the target points' positions to maintain a safe distance between the gripper fingers and the object. The offset translates the target points in the direction
of the gripper base (see the arrow in Fig. \ref{fig:main}b). Reducing this offset brings the gripper closer to the object. This is done in the last phase of grasping (as detailed in  Section \ref{gripcontrol}).

\subsubsection{Auxiliary Controller Module} 

This module implements collision avoidance by realizing a virtual repulsive region around the human hand and the object. In combination with the Main Controller Module, the robot can negotiate a collision-free path while reaching the target. The repulsive action is defined by several spherical repulsive regions, each implemented by a repulsive spring with a Gaussian-like force magnitude profile.
Given the energy $E=k_r\sigma_r^2e^{(\frac{-|\textbf{p}|^2}{2\sigma_r^2})}$, 
\begin{equation}
   \textbf{F}_r = -|\textbf{p}_r|k_re^{(\frac{-|\textbf{p}|^2}{2\sigma_r^2})}
    \label{eq:repulsive}
\end{equation}
where $k_r > 0$ is the stiffness, $\sigma_r$ is the distance at which the repulsive region applies its maximum repulsion force, and $\textbf{p}_r$ is 
the displacement vector between the center of the repulsive region and the gripper fingers. If we define $k_r = \frac{F_{r_{max}}}{\sigma_r}e^{0.5}$, then the parameter $F_{r_{max}}$ represents the maximum force the repulsive spring can applies. In particular, the maximum force $F_{r_{max}}$ is generated at the distance $|\textbf{p}_r| = \sigma_r$. 

The number and placement of these regions depend on the task. In our experimental setup, the robot always approaches the object/human from the top or the front. Therefore, we use a single spherical repulsive region with its center always at a distance $\gamma$ ($\gamma = 23$ cm, determined experimentally) from the hand, as shown in Fig. \ref{fig:main}b, to protect the space in front of the object/hand. In practice, other scenarios could involve more complex interactions that require additional repulsive regions to ensure human and environmental safety. Ideally, real-time mapping of the environment and humans would guide the placement of these regions.

\subsection{Virtual Model Controller Implementation}

In virtual model control \cite{pratt2001virtual,larby2024optimal}, each component of the virtual mechanisms (i.e., springs and dampers) applies a force $\textbf{F}_i$ at a point $\textbf{p}_i$ on the robot body to realize the desired behavior. For a robot arm whose revolute joints have coordinates $\textbf{q}$, the point $\textbf{p}_i$ can be expressed as a function of the joint coordinates $\textbf{p}_i = \textbf{h}_{\textbf{p}_i}(\textbf{q})$, and the forces from all the virtual components are translated into joint torques $\boldsymbol{\tau}$ by 
\begin{equation}
    \boldsymbol{\tau} = \sum_{i=1}^{n} \textbf{J}_{\textbf{p}_i}^T(\textbf{q})\textbf{F}_i,
\end{equation}
where $\textbf{J}_{\textbf{p}_i}^T(\textbf{q})$ is the transpose of the Jacobian matrix $\textbf{J}_{\textbf{p}_i}(\textbf{q}) = \frac{\partial \textbf{h}_{\textbf{p}_i}(\textbf{q})}{\partial \textbf{q}}$. We refer the reader to \cite{larby2024optimal} for a detailed
characterization of virtual model control. 

We use the software package \texttt{VMRobotControl.jl} package \cite{VMRobotControl} for implementing the controller on a Franka Emika Panda robot arm equipped with the Franka Hand gripper. The package simplifies the design of virtual model control mechanisms and components and facilitates their integration with robot manipulators. Furthermore, it provides a simulation environment that facilitates the evaluation of the virtual mechanism behavior and its interaction with the robot, as well as rapid iteration of controller parameters.
We utilize the \texttt{VMRobotControl.jl} simulation environment to design and tune the proposed virtual model controller, followed by fine-tuning on the real robot through trial and error. The length of 
the rigid links attached to gripper and object 
was set at $45$ cm upon experimentally observing the robot's reaction to rotating and moving the hand. 
Larger lengths led to instability and response issues, such as gripper oscillations following small hand rotations (due to the finite bandwidth of the joint actuators). Shorter lengths made the robot less reactive. All remaining parameters (stiffness and damping) are summarized in Fig. \ref{fig:vmcparams}.

\subsection{Gripper Control Module}
\label{gripcontrol}
The final grasping and handover action is handled by a standalone module to mitigate the lack of torque/force control ability of the hardware adopted in our experiment (Franka Hand gripper). When the Gripper Control Module is activated, the gripper moves closer to the object by reducing the offset $\alpha$ to zero, followed by a closure of the gripper fingers to complete the task. To activate this module, two conditions must be satisfied:
(i) the hand velocity magnitude is low $|v_h|<5$ cm/s and (ii) all the distances between paired gripper-target points must be below a certain value $d<10$ cm. Grasping is then initiated after at least one second if $d<5$ cm. All values were fitted experimentally.

This module runs in parallel with the interaction layer. Therefore, if the hand moves when the grasping action is not complete, the module is deactivated by resetting the controller offset $\alpha$ to $10$ cm and reopening the gripper, which allows the interaction layer to continue the handover even if the object is moved when the robot is closing its fingers.

\begin{figure}[t!]
  \vspace{2.2mm} 
  \centering
  \includegraphics[width=0.95\linewidth]{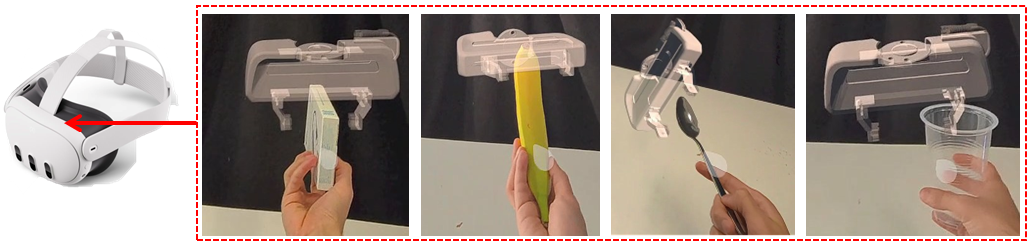}
  \caption{Objects used in the experiments and their grasping pose as seen through the augmented reality interface: (left to right) a cardboard box, a banana, a spoon, and a plastic cup.}
  \label{fig:poses}
\end{figure}

\section{Bidirectional Communication using Augmented Reality}

We utilize augmented reality to facilitate human-robot bidirectional communication, given its potential to enhance human-robot interaction \cite{suzuki2022augmented, costa2022augmented}. We rely on the Meta Quest 3 headset to harness its hand-tracking capability and leverage it as a communication medium. From the Quest 3, the robot receives the measured hand palm pose, whereas the human sees a visualization of the expected grasping pose. This enables the human to modify the hand pose and therefore the object pose before the robot attempts to grasp it, thereby improving the likelihood of successful handover.

We use the Meta XR All-in-One SDK and the Unity game engine for programming the Quest 3 and communicating with the robot. From the SDK, we use the available hand-tracking capability, which provides a hand skeleton with the pose of 26 points. Additionally, the Mixed Reality Utility Kit (MRUK) API is used to scan the room and calibrate the robot location in the Quest 3 coordinate frame. The latter is only performed once by placing a \texttt{MRUKAnchor} on the robot origin. Then, the MRUK tool loads the room scanning, including detection of any \texttt{MRUKAnchor} in the room, whenever the device is launched in the saved area.

As the Quest 3 does not provide access for object detection and tracking, we construct the controller target points by considering low-pass filtered readings of the hand palm pose with the dimensions of four known objects (see Fig. \ref{fig:poses}). These objects are chosen to accommodate the inaccuracies and limitations of the Quest 3 hand-tracking, which is heavily influenced by hand occlusion, while also being topologically similar to objects tested in previous object handover works. Finally, we utilize a Kalman filter to estimate the hand velocity that is used in the Gripper Control Module.

\begin{table}[b!]
\setlength{\tabcolsep}{4pt}
\centering
\renewcommand{\arraystretch}{1.1}
\caption{Performance metrics. The figure shows the points used to calculate the distance-related metrics.}
\begin{minipage}[c]{0.5\linewidth}
\scriptsize
\begin{tabular}{c|c}
\Xhline{2\arrayrulewidth}
Metric & Definition \\ \hline
$t_a$ (s) & \begin{tabular}[c]{@{}c@{}} Approach time from when the robot starts \\ moving till the gripper closes on the object\end{tabular} \\ \hline
SR ($\%$)& Success rate of the object handover \\ \hline
$d_i$ (cm)& Initial distance between the robot and the object \\ \hline
$L_r$ (cm) & Total path length covered by the robot motion \\ \hline
$L_o$ (cm) & Total path length covered by the object motion \\ \hline
$e_d$ (cm) & \begin{tabular}[c]{@{}c@{}}Distance error between the gripper points \\ and the target points during the grasping\end{tabular} \\ \hline
$\theta_i$ (deg) & \begin{tabular}[c]{@{}c@{}}Initial angle difference between \\ the robot and the initial expected pose \end{tabular} \\ \hline
$\theta_r$ (deg) & \begin{tabular}[c]{@{}c@{}}Difference of the robot angle \\ between the start and end of experiment\end{tabular} \\ \hline
$\theta_o$ (deg) & \begin{tabular}[c]{@{}c@{}}Difference of the object angle \\ between the start and end of experiment\end{tabular} \\ \hline
$e_\theta$ (deg) & \begin{tabular}[c]{@{}c@{}}Error in angle between the gripper points \\ and the target points during the grasping\end{tabular} \\ \Xhline{2\arrayrulewidth}
\end{tabular}
\label{tab:metrics}
\end{minipage}
\hspace{2.1cm}
\begin{minipage}[c]{0.23\linewidth}
\centering
\includegraphics[width=\linewidth]{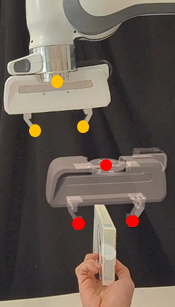} 

\end{minipage}
\end{table}

\section{Resilient Human-to-Robot Object Handovers}

\begin{figure*}[t!]
  \vspace{2.2mm} 

  \centering

  \includegraphics[width=0.99\linewidth]{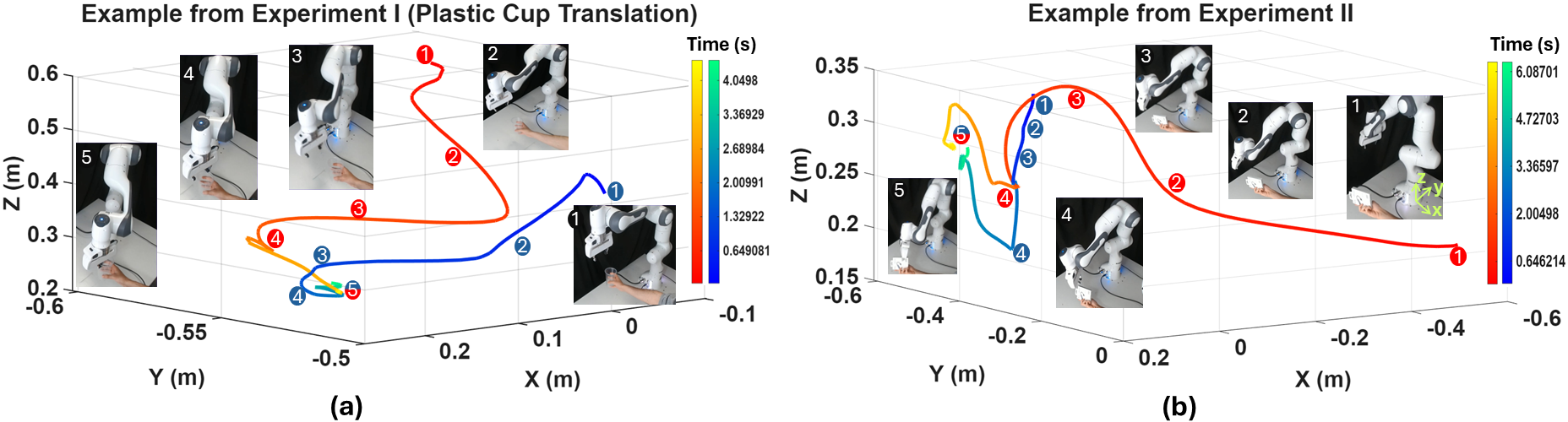}
  \caption{Examples of trajectories from the gripper right finger (\textcolor{red}{red}) and its corresponding target point (\textcolor{blue}{blue}) with temporal markers of recorded snapshots, where (a) is a plastic cup translation example from Experiment I with the object being moved after the robot starts moving and (b) is an example from Experiment II highlighting the effect of the Auxiliary Controller Module with a spherical repulsive region in front of the object pushing the gripper above the object between markers 2 and 3 (instead of continuing through a straight line). Robot coordinate system used as the reference frame is shown in photo 1 of (b). Full experimental runs of the presented trajectories are shown in the Supplementary Video.}
  \label{fig:trajectories}
\end{figure*}

\begin{table*}[ht!]
\centering
\footnotesize
\caption{Mean (standard deviation) of the proposed metrics for Experiment I}
\begin{tabular}{ccccccccc}
\Xhline{2\arrayrulewidth} 
\multirow{2}{*}{} & \multicolumn{2}{c}{Cardboard Box} & \multicolumn{2}{c}{Banana} & \multicolumn{2}{c}{Spoon} & \multicolumn{2}{c}{Plastic Cup} \\
\cline{2-9}
 & Rotation & Translation & Rotation & Translation & Rotation & Translation & Rotation & Translation \\
\hline
$t_a$ & 3.38 (0.34) & 4.79 (0.54) & 3.14 (0.37) & 4.14 (0.58) & 3.64 (0.91) & 4.30 (0.39) & 3.80 (0.52) & 4.76 (0.86) \\
\hline
SR & 100\% & 100\% & 100\% & 100\% & 95\% & 100\% & 100\% & 100\% \\
\hline
$d_i$ & 30.88 (3.42) & 22.67 (4.51) & 26.63 (2.47) & 19.76 (2.79) & 27.05 (2.63) & 25.15 (4.46) & 33.69 (3.46) & 26.90 (7.62) \\
\hline 
$L_r$ & 47.43 (4.70) & 68.55 (14.32) & 49.44 (4.37) & 63.97 (12.26) & 46.66 (5.81) & 66.21 (11.30)  & 50.22 (6.59) & 73.62 (19.52) \\
\hline 
$L_o$ & 15.97 (4.14) & 52.84 (18.68) & 20.88 (4.09) & 42.56 (12.67) & 19.99 (3.46) & 39.27 (12.33) & 20.49 (6.70) & 53.43 (19.45) \\
\hline 
$e_d$ & 2.02 (1.02) & 1.79 (0.44) & 3.88 (0.47) & 3.06 (0.78) & 3.44 (1.21) & 3.42 (0.88) & 2.63 (0.74) & 1.95 (0.71) \\
\hline
$\theta_i$ & 16.54 (3.78) & 11.86 (4.68) & 94.27 (6.13) & 98.77 (4.17) & 86.41 (13.61) & 83.57 (4.41) & 12.62 (5.62) & 13.01 (4.28) \\
\hline
$\theta_r$ & 38.43 (7.02) & 20.87 (5.90) & 107.73 (7.02) & 97.73 (5.25) & 103.98 (18.49) & 90.16 (6.80) & 43.85 (10.89) & 14.89 (4.28) \\
\hline
$\theta_o$ & 30.88 (6.45) & 12.70 (5.49) & 43.29 (10.99) & 16.36 (8.07) & 52.20 (24.41) & 13.99 (5.48) & 40.44 (11.71) & 9.86 (3.68) \\
\hline
$e_\theta$& 2.46 (1.07) & 2.51 (1.33) & 3.69 (1.81) & 3.04 (1.26) & 3.46 (2.13) & 5.05 (1.57) & 1.97 (0.88) & 1.96 (0.82) \\
\Xhline{2\arrayrulewidth} 
\end{tabular}
\label{tab:experimentI}
\end{table*}

\subsection{Experimental Protocol}
We evaluated our controller in two experiments that cover a significant 
set of realistic scenarios mainly focusing on variations in object poses that can occur during the handovers.
Experiment I considers handover with 
different objects that are also moved by the user (translation and/or rotation) after the robot starts
to reach them. This
experiment evaluates the robot's resilience to changes in the expected grasping pose due to variable sizes and shapes of the objects and object movements, such that the robot can continue the handover process in the presence of unpredictable human hand motion.
Experiment II looks into randomized robot starting poses and also extends the set of potential object movements to include complex sequences of movements at different times.
These experiments are supplemented with a user study to evaluate the user experience in interacting with the robot, and the feasibility of the augmented reality for communication.

In each experimental run, a person holds an object in the hand in a fixed, predetermined position compatible with the grasping capability of a two-finger gripper (see Fig. \ref{fig:poses} for an example).
The robot's external enabling device is pressed to initiate the robot's movement. Subsequently, the person moves object to induce a change in its grasping pose that the robot needs to adapt to. Each experimental run ends when the gripper closes its fingers on the object, as we do not consider the post-handover phase. The handover is considered successful if the object is grasped successfully. We exclude from the analysis any experimental run that encountered communication loss in the system or corrupted hand-tracking algorithm of the Quest 3, regardless of its success. Hence, failures are counted when the robot cannot reach the final pose and fails to grasp the object (e.g., closing the fingers before reaching the final pose).

\begin{table}[t!]
\caption{Mean $t_a$ of our method against similar rotation (R) and translation (T) tests reported in previous works. All methods achieved 100\% SR unless stated otherwise{$^\dagger$}.}
\label{tab:comparison}
\centering

\scriptsize
\setlength{\tabcolsep}{5.75pt}
\begin{tabular}{ccccccccc}
\Xhline{2\arrayrulewidth}

\multicolumn{1}{c}{} & \multicolumn{2}{c}{Cardboard Box} & \multicolumn{2}{c}{Banana} & \multicolumn{2}{c}{Spoon} & \multicolumn{2}{c}{Cup/Mug} \\ \hline
\multicolumn{1}{c}{} & \multirow{1}{*}{R} & \multirow{1}{*}{T} & \multirow{1}{*}{R} & \multirow{1}{*}{T} & \multirow{1}{*}{R} & \multirow{1}{*}{T} & \multirow{1}{*}{R} & \multirow{1}{*}{T} \\ \hline
\multicolumn{1}{c}{Ours} & \multicolumn{1}{c}{\textbf{3.38}} & \multicolumn{1}{c}{\textbf{4.79}} & \multicolumn{1}{c}{\textbf{3.14}} & \multicolumn{1}{c}{\textbf{4.14}} & \multicolumn{1}{c}{\textbf{3.64}$^\dagger$} & \multicolumn{1}{c}{\textbf{4.30}} & \multicolumn{1}{c}{\textbf{3.80}} & \multicolumn{1}{c}{\textbf{4.76}} \\ \hline
\multicolumn{1}{c}{\cite{zhang2023flexible}} & \multicolumn{1}{c}{3.43} & \multicolumn{1}{c}{5.87} & \multicolumn{1}{c}{4.43} & \multicolumn{1}{c}{5.65} & \multicolumn{1}{c}{4.29$^\dagger$} & \multicolumn{1}{c}{5.86$^\dagger$} & \multicolumn{1}{c}{5.29} & \multicolumn{1}{c}{5.77} \\ \hline
\multicolumn{1}{c}{\cite{yang2021reactive}} & \multicolumn{1}{c}{8.23} & \multicolumn{1}{c}{-} & \multicolumn{1}{c}{16.34} & \multicolumn{1}{c}{-} & \multicolumn{1}{c}{-} & \multicolumn{1}{c}{-} & \multicolumn{1}{c}{-} & \multicolumn{1}{c}{-} \\ \hline
\multicolumn{1}{c}{\cite{yang2022model}*} & \multicolumn{1}{c}{5$^\dagger$**} & \multicolumn{1}{c}{-} & \multicolumn{1}{c}{5$^\dagger$**} & \multicolumn{1}{c}{-} & \multicolumn{1}{c}{-}  & \multicolumn{1}{c}{-}  & \multicolumn{1}{c}{-}  & \multicolumn{1}{c}{-} \\
\Xhline{2\arrayrulewidth}
\\
\end{tabular}
\parbox{\linewidth}{\scriptsize
\vspace{-7pt}
\raggedright
$^\dagger$ SR $<100\%$: $95\%$ in Ours-Spoon-R, $75\%$ in \cite{zhang2023flexible}-Spoon-R/T, $50\%$ in \cite{yang2022model}. \\ * The fastest variation (MPC-GS) is reported here out of three tested variations. \\ ** $t_a$ and SR in \cite{yang2022model} were reported as the average across three different objects.
}
\end{table}

\subsection{Experimental Metrics}
Existing works and benchmarks metrics are mostly tailored for static object handovers and focus on elaborate assessment of the sensory components \cite{sanchez2020benchmark}. Researchers tend to rely on the approach time and success rates when assessing the resilience of their methods. In contrast, we acquire a comprehensive set of metrics targeted towards evaluating the handover resilience, summarized in Table \ref{tab:metrics}. Each experimental run begins when the external activation device is pressed (i.e., the robot starts moving) and ends when the gripper fingers close. 

Distance-related metrics ($d_i, L_r, L_o, e_d$) are all calculated from the points illustrated in Table \ref{tab:metrics}. 
Yellow-red point pairs are used for $d_i$ and $e_d$.
Red points are used for $L_o$, whereas yellow points are used for $L_r$. The largest of the three values is reported in the results.
Angles are calculated based on the axis-angle representation formed by the relative rotation matrix between the two relevant rotation matrices. For example, in the case of $\theta_r$, the two rotation matrices are formed by the three virtual model control points attached to the robot at its initial pose and at its final pose. 

The approach time is correlated to the distance or angle traveled by the robot, which influences the duration of a given experiment. If the object moves a short distance, the experiment will be naturally shorter and vice-versa.
Likewise, the path length covered by the robot is correlated to the movement of the object and to the initial distance and angle between the robot and the object. 

The errors in distance and angle between the target points and the gripper points provide insights into the controller's performance under current settings, and its suitability for other objects (e.g., a tiny object may require stronger spring stiffness to reduce errors, which may negatively affect other aspects of the system).

\subsection{Experiment I: Object Translation and Rotation}

\begin{table*}[t!]
\vspace{2.2mm} 

\caption{Mean (standard deviation) of the metrics for Experiment II}
\centering
\scriptsize

\begin{tabular}{c c c c c c c c c c}
\Xhline{2\arrayrulewidth}
 $t_a$ & SR & $d_i$ & $L_r$ & $L_o$ & $e_d$ & $\theta_i$ & $\theta_r$ & $\theta_o$ & $e_\theta$ \\ \hline
6.79 (0.91) & 100\% & 61.89 (8.16) & 122.31 (16.79) & 55.50 (18.07) & 2.32 (0.75) & 69.19 (30.27) & 78.52 (29.99) & 31.55 (15.24) & 2.97 (1.46) \\ 
\Xhline{2\arrayrulewidth}
\end{tabular}
\label{tab:experimentII}
\end{table*}

This experiment adopts the one-motion experimental protocol in \cite{yang2021reactive, zhang2023flexible, yang2022model}. Four objects requiring different grasping poses (shown in Fig. \ref{fig:poses}) are moved once randomly, either by translation or rotation, immediately after the robot starts moving. The robot always starts from the same initial pose. We perform 20 randomized experimental runs for each object and each motion type (rotation or translation). 

Our proposed controller achieves high success rates across all tests, as summarized in Table \ref{tab:experimentI}. The mean approach time was less than $4$s when the object was rotated and less than $5$s  for translations, which was faster than the approach time found in \cite{zhang2023flexible, yang2021reactive, yang2022model} while achieving similar or higher success rates, as reported in Table \ref{tab:comparison}. This demonstrates the proposed interaction layer's ability to quickly adapt and handle the tested pose changes. The final distance and angle errors varied among objects, likely caused by poses that are intrinsically more difficult and less reachable given the robot's kinematics. This is the case of both banana and spoon grasping (large initial angle $\theta_i$ in addition to the induced motion). 

Fig. \ref{fig:trajectories}a provides an
illustration of the trajectories of the gripper right finger and its paired 
target point associated to a plastic cup.
This is a translation experimental run, with markers 1 to 5 corresponding to snapshots from the recorded video. Starting at marker 1, the object was moved away from the robot, which resulted in the extension of the virtual springs and dampers causing the robot to closely follow the target point. At marker 4, the hand was nearly static
and there was close proximity between the object and the gripper. This activated the Gripper Control Module. The robot executed a final approach 
until grasping was achieved, at marker 5.

\subsection{Experiment II: Random Starting Pose and Movements}

In this experiment, we complete 20 experimental runs that demonstrate resilience against random robot starting poses and object motion. The object is moved when the robot is near it and after the Gripper Control Module is activated. The experiment is limited to the cardboard box object as the analysis focuses on the random robot starting pose and object motion. Nonetheless, we anticipate that the results of this experiment generalize to other objects, since Experiment I already demonstrated strong resilience to grasping poses across different objects.

Successful robot grasping is consistently achieved in all experimental runs. In comparison to the previous experiment, both $e_d$ and $e_\theta$ were higher, albeit without a strong effect on the success rate (see Table \ref{tab:experimentII}). 
When the gripper starting pose was below the human hand, the Auxiliary Controller Module enabled the robot to conveniently approach the object from the top, as shown in Fig. \ref{fig:trajectories}b. The robot was pushed upwards between markers 2 and 3. This prevented a potential collision with the object. At marker 4, the object was pulled away by the user right after the Gripper Control Module triggered an attempt to grasp. The reactive behavior granted by the  interaction layer allowed the robot to move towards the new position of the object, 
achieving a successful grasping at marker 5. 

\subsection{Further considerations and limitations}

The results strongly support our hypothesis that an interaction layer based on virtual model control enables resilient human-to-robot object handovers. Nonetheless, the proposed controller falls short in certain cases. Currently, the robot struggles to continue the handover efficiently when its joints are near their limits and when the object is too close to the robot body. Such cases can be mitigated by designing additional virtual components such as joint-level repulsive springs and localized repulsive springs regions near the robot body. These components can also be designed to increase their forces if the robot is stuck for a period of time.

Another limitation in the proposed interaction layer is that it is designed for two-fingered grippers. Since the position of each finger is controlled by a single virtual rigid link and virtual components connecting one pair of gripper-target points, extensions towards multi-fingered grippers appear to be straightforward, simply by adding more virtual links, more gripper-target pair per finger, and further parameter tuning. However, the specific action of grasping would need to address the complexity of multi-finger coordination. This requires further research.
This study serves as a proof of concept of the interaction layer and its associated virtual model controller. Therefore, we restricted the experiments to a two-fingered gripper and avoided actions that could stall the robot. Addressing these limitations will be considered in future works to improve the generalization of our method. 

\section{User Study}

\subsection{Experimental Conditions}

We conducted a within-subject user study to evaluate the performance of our method across different users and to understand user preferences in augmented reality-based human-to-robot object handovers. Four conditions were considered with two different robot profiles and the presence and absence of augmented reality visuals. Condition \textbf{A} had the visuals turned on and the same controller parameters used in the previous experiments, resulting in an authoritative, resilient robot that can handle the handover by itself. Condition \textbf{B} had the same authoritative robot but without the visuals. Condition \textbf{C} had the visuals turned on, but with a slower, less resilient robot that required the human to bring the object closer to the robot, resulting in a more cooperative nature of the handover. Finally, Condition \textbf{D} kept the cooperative robot profile without the visuals. 
The cooperative robot profile was enabled by setting $F_{s_{max_2}}$ to zero.

\begin{figure*}[ht!]
  \vspace{2.2mm} 
  \centering
  \includegraphics[width=0.97\linewidth]{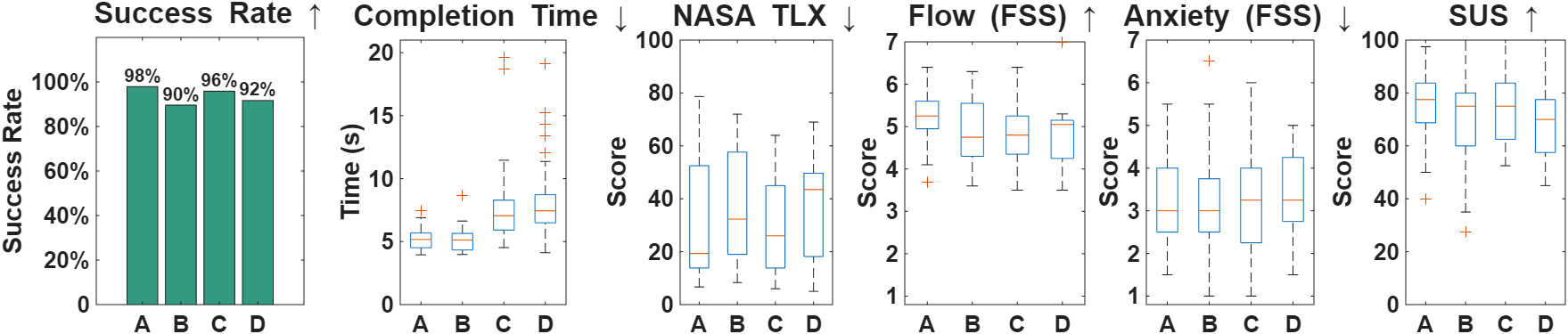}
  \caption{Summary of results from the user study experiments highlighting the success rate and completion time, as well as scores of the three surveys used for assessment (NASA TLX, FSS, and SUS).}
  \label{fig:userstudy}
\end{figure*}

\subsection{Participants and Experimental Protocol}

We recruited 16 participants (8 females and 8 males) through opportunity sampling. They were between 23 and 34 years old ($M=28.06$, $SD=3.17$), and none of them had prior experience with augmented reality; however, 13 participants reported varying levels of earlier exposure to virtual reality. Moreover, 9 participants had experience with robotics, with 8 reporting extensive exposure, while the remaining 7 had no prior exposure to robotics.

We asked the participants to move the cardboard box object in a single translation motion stroke after the robot started moving. In Conditions \textbf{A} \& \textbf{B}, participants were told to stop moving after that and wait for the robot to grasp the object, in contrast to Conditions \textbf{C} \& \textbf{D}, where they were asked to wait for the robot to approach them and move the object to it afterwards. The order of the conditions was randomized so that each participant performed a unique sequence of conditions.

Each participant was shown one demonstration of each condition, accompanied by a translation motion of the object, before starting the experiments. In addition, a five-minute familiarization phase was provided with Quest 3 to understand how the visuals work and align the object as necessary (without operating the robot). Each participant was asked to perform three experimental runs per condition. Participants were allowed an additional run in case they did not follow the instructions in the first run (e.g., not moving the object or moving out of the robot workspace). This additional attempt replaces the first attempt regardless of its success status.

 After each condition, we asked the participants to fill out three surveys: \textit{NASA Task Cognitive Load} (TLX) \cite{hart_1988_development}, \textit{Flow Short Scale} (FSS) \cite{engeser_flow_2Sanfeliu008}, and \textit{System Usability Scale} (SUS) \cite{Brooke1996}. At the end of the experiment, each participant was involved in a semi-structured discussion to better understand their preferences and feedback.

\subsection{Experimental Results}

We show the success rate, completion time of successful handovers, and survey scores in Fig. \ref{fig:userstudy}. Specifically, Condition \textbf{A}, which adopted the proposed virtual model control interaction layer and augmented reality approach, led to a high success rate (98\%) and the lowest completion time ($M = 4.9$s), which was close to the reported translation approach (see Table \ref{tab:experimentI}). This indicates the effectiveness of deploying the proposed method for new users.

Overall, all four conditions (\textbf{A}--\textbf{D}) delivered high success rates with reasonable completion times. Nonetheless, conditions utilizing augmented reality visuals (\textbf{A} and \textbf{C}) resulted in higher usability and lower task load, with the authoritative robot (\textbf{A}) providing better results than the cooperative robot (\textbf{C}). Moreover, FSS scores indicated inconsistency of augmented reality impact across the two robot profiles. With the authoritative robot, participants were more engaged and experienced lower anxiety levels in the presence of augmented reality visuals, in contrast to the cooperative robot profile, leading to opposite observations.

\subsection{Statistical Analysis}

The Shapiro-Wilk test revealed that only completion time under Condition \textbf{B} ($p = 0.0062$) and the NASA TLX under Condition \textbf{A} ($p = 0.0098$) were not normally distributed. Therefore, we used a non-parametric Friedman test to analyze the mean of the successful trials and Kendall's W value to compute the effect size, which showed a statistically significant difference ($\chi^{2}(3)=32.4750,p\leq0.0001,W=0.677$) for the completion time and a non-significant result for NASA TLX ($\chi^{2}(3)=1.425,p=0.699,W=0.030$).

Post hoc analysis of completion time using the Nemenyi test showed a statistically significant difference between conditions \textbf{A} and \textbf{C} (p = 0.0021), conditions \textbf{A} and \textbf{D} ($p\leq0.0001$), conditions \textbf{B} and \textbf{C} ($p = 0.0035$), and conditions \textbf{B} and \textbf{D} ($p\leq0.0001$). This difference was expected as the cooperative conditions were slower and required a longer time to complete the task.

To analyze the remaining data, we used a one-way repeated measures ANOVA test and the partial eta-squared value for the effect size. The results showed a non-significant difference for flow ($F(3,45)=1.026,p= 0.3899,\eta_p^2=0.064$), and anxiety ($F(3,45)=0.327,p= 0.805,\eta_p^2=0.02$) measured using the FSS questionnaire, and SUS survey scores ($F(3,45)=1.322,p= 0.279,\eta_p^2=0.081$). These results reflect the slight variation in preferences observed during the semi-structured discussion. 

To address potential bias, we conducted a Spearman’s rank correlation analysis between participants’ prior experience and their performance across all metrics. Specifically, we compared previous robotics experience against the means of the authoritative ($\textbf{A+B}$) and cooperative ($\textbf{C+D}$) conditions, and experience with immersive interfaces against the augmented reality-activated ($\textbf{A+C}$) conditions. No significant correlations were found ($p > 0.05$ in all tests), suggesting that prior expertise did not substantially influence the results.

\subsection{Participants' Feedback and Comments}
Our semi-structured discussion revealed that 7 participants (43.75\%) preferred \textbf{A}, 5 (31.25\%) preferred \textbf{B}, 3 (18.75\%) preferred \textbf{C}, with only 1 preferring \textbf{D}. The 12 participants who preferred the authoritative robot profile mentioned liking the robot's responsiveness and handling most of the task effort. In contrast, those who preferred the cooperative robot liked being involved when giving the object and compensating for any inaccuracies coming from the robot. 

Regarding augmented reality visuals, the 10 participants who preferred them mentioned that knowing the expected position of the robot gave them confidence in the task. In contrast, the remaining six participants found the visuals too distracting from the operation of the real robot, aligning with prior findings \cite{newbury2022visualizing}. Additionally, a handful of participants expressed confusion about when to give the object to the robot with the cooperative robot profile, which hints at a need for better communication of the robot's current state.

\section{Conclusion and Future Works}

We have proposed an interaction layer for resilient human-to-robot handovers based on Virtual Model Control and Augmented Reality (AR) bidirectional communication. The interaction layer is designed as a virtual, compliant mechanical interconnection between the robotic gripper and a generated grasp pose. This enables resilient robot behaviors that adapt and react to object or human motion. The approach demonstrates robustness to various disturbances while maintaining fast approach times (less than 4s for rotations and less than 5s for translations) and near-perfect success rates. These results indicate that the virtual interaction layer effectively guarantees resilient grasping without the need for continuous trajectory re-planning.

Our user study focused on analyzing preferences regarding robot behavior and augmented reality. We gathered valuable insights into how AR-based visual information complements fast-resilient versus slow-cooperative robot behavior profiles. The study demonstrates a strong preference for the fast-resilient profile and a consistently positive response to the AR visuals.

These studies highlight the need for an adaptive interaction layer capable of offering a personalized experience, which will be a focus of our future research. The effectiveness of this approach motivates the development of a full object-handover pipeline, which will require validation on a wider object set and with various grippers. This will involve the integration of robust hand–object segmentation and grasp-generation methods, as well as improving system safety under sensor noise and during close-proximity human–robot interactions.

\section*{Acknowledgments}

The Cambridge University Engineering Department Ethics Committee approved the user study and its procedure. 

\bibliographystyle{IEEEtran}
\bibliography{references}

@article{ortenzi2021object,
  title={Object handovers: a review for robotics},
  author={Ortenzi, Valerio and others},
  journal={IEEE Trans. Robot.},
  volume={37},
  number={6},
  pages={1855--1873},
  year={2021},
  publisher={IEEE}
}

@article{duan2024human,
  title={Human-robot object handover: Recent progress and future direction},
  author={Duan, Haonan and others},
  journal={Biomim. Intell. Robot.},
  pages={100145},
  year={2024},
  publisher={Elsevier}
}

@article{rosenberger2020object,
  title={Object-independent human-to-robot handovers using real time robotic vision},
  author={Rosenberger, Patrick and others},
  journal={IEEE Robot. Autom. Lett.},
  volume={6},
  number={1},
  pages={17--23},
  year={2020},
  publisher={IEEE}
}

@article{engeser_flow_2Sanfeliu008,
	title = {Flow, performance and moderators of challenge-skill balance},
	volume = {32},
	issn = {1573-6644},
	doi = {10.1007/s11031-008-9102-4},
	language = {en},
	number = {3},
	urldate = {2023-07-03},
	journal = {Motiv. Emot.},
	author = {Engeser, Stefan and Rheinberg, Falko},
	month = sep,
	year = {2008},
	pages = {158--172},
}

@article{hart_1988_development,
  author = {Hart, Sandra and Staveland, Lowell},
  pages = {139--183},
  title = {{Development of NASA-TLX (Task Load Index): Results of Empirical and Theoretical Research}},
  year = {1988},
  journal = {Advances in Psychology}
}

@article{Brooke1996,
author = {Brooke, John},
year = {1996},
month = {November},
pages = {189-194},
title = {{SUS: A 'Quick and Dirty' Usability Scale}},
volume = {189},
booktitle = {{Usability Evaluation in Industry}}
}

@INPROCEEDINGS{10802282,
  author={Tadeja, Sławomir K. and Zhou, Tianye and Capponi, Matteo and Walas, Krzysztof and Bohné, Thomas and Forni, Fulvio},
  booktitle={2024 IEEE/RSJ Int. Conf. Intell. Robots Syst. (IROS)}, 
  title={{Using Augmented Reality in Human-Robot Assembly: A Comparative Study of Eye-Gaze and Hand-Ray Pointing Methods}}, 
  year={2024},
  volume={},
  number={},
  pages={8786-8793},
  doi={10.1109/IROS58592.2024.10802282}}

@article{castellani2024soft,
  title={Soft Human-Robot Handover using a Vision-Based Pipeline},
  author={Castellani, Chiara and others},
  journal={IEEE Robot. Autom. Lett.},
  year={2024},
  publisher={IEEE}
}

@inproceedings{newbury2022visualizing,
  title={Visualizing robot intent for object handovers with augmented reality},
  author={Newbury, Rhys and others},
  booktitle={2022 31st IEEE Inter. Conf. on Robot and Human Inter. Comm. (RO-MAN)},
  pages={1264--1270},
  year={2022},
  organization={IEEE}
}

@article{costa2022augmented,
  title={Augmented reality for human--robot collaboration and cooperation in industrial applications: A systematic literature review},
  author={Costa, Gabriel de Moura and Petry, Marcelo Roberto and Moreira, Ant{\'o}nio Paulo},
  journal={Sensors},
  volume={22},
  number={7},
  pages={2725},
  year={2022},
  publisher={MDPI}
}

@inproceedings{suzuki2022augmented,
  title={Augmented reality and robotics: {A} survey and taxonomy for ar-enhanced human-robot interaction and robotic interfaces},
  author={Suzuki, Ryo and others},
  booktitle={2022 ACM Conf. Hum. Factors Comput. Syst. },
  pages={1--33}
}

@article{larby2025collaborative,
  title={{Collaborative Drill Alignment in Surgical Robotics}},
  author={Larby, Daniel and Kershaw, Joshua and Allen, Matthew and Forni, Fulvio},
  journal={arXiv:2503.05791},
  year={2025}
}

@article{wang2021predicting,
  title={Predicting human intentions in human--robot hand-over tasks through multimodal learning},
  author={Wang, Weitian and Li, Rui and Chen, Yi and Sun, Yi and Jia, Yunyi},
  journal={IEEE Trans. Autom. Sci. Eng.},
  volume={19},
  number={3},
  pages={2339--2353},
  year={2021},
  publisher={IEEE}
}

@article{sanchez2020benchmark,
  title={Benchmark for human-to-robot handovers of unseen containers with unknown filling},
  author={Sanchez-Matilla, Ricardo and others},
  journal={IEEE Robot. Autom. Lett.},
  volume={5},
  number={2},
  pages={1642--1649},
  year={2020},
  publisher={IEEE}
}

@inproceedings{yang2022model,
  title={Model predictive control for fluid human-to-robot handovers},
  author={Yang et al., Wei},
  booktitle={2022IEEE Int. Conf. Robot. Autom. (ICRA)},
  pages={6956--6962},
  year={2022},
  organization={IEEE}
}

@article{liu2021object,
  title={Object transfer point predicting based on human comfort model for human-robot handover},
  author={Liu, Dong and Wang, Xianwei and Cong, Ming and Du, Yu and Zou, Qiang and Zhang, Xiaomin},
  journal={IEEE Trans. Instrum. Meas.},
  volume={70},
  pages={1--11},
  year={2021},
  publisher={IEEE}
}

@inproceedings{zhang2024virtual,
  title={Virtual model control for compliant reaching under uncertainties},
  author={Zhang, Yi and Larby, Daniel and Iida, Fumiya and Forni, Fulvio},
  booktitle={2024 IEEE/RSJ Int. Conf. Intell. Robots Syst. (IROS)},
  pages={795--801},
  year={2024},
  organization={IEEE}
}

@article{chen2020virtual,
  title={Virtual model control for quadruped robots},
  author={Chen, Guangrong and Guo, Sheng and Hou, Bowen and Wang, Junzheng},
  journal={IEEE Access},
  volume={8},
  pages={140736--140751},
  year={2020},
  publisher={IEEE}
}

@inproceedings{winkler2014path,
  title={Path planning with force-based foothold adaptation and virtual model control for torque controlled quadruped robots},
  author={Winkler, Alexander and others},
  booktitle={2014 IEEE Int. Conf. Robot. Autom. (ICRA)},
  pages={6476--6482},
}

@inproceedings{zhang2023flexible,
  title={Flexible handover with real-time robust dynamic grasp trajectory generation},
  author={Zhang, Gu and Fang, Hao-Shu and Fang, Hongjie and Lu, Cewu},
  booktitle={2023 IEEE/RSJ Int. Conf. Intell. Robots Syst. (IROS)},
  pages={3192--3199},
}

@inproceedings{wang2024genh2r,
  title={GenH2R: Learning Generalizable Human-to-Robot Handover via Scalable Simulation Demonstration and Imitation},
  author={Wang, Zifan and others},
  booktitle={ IEEE/CVF Comput. Soc. Conf. Comput. Vis. Pattern Recognit.},
  pages={16362--16372},
  year={2024}
}

@article{pratt2001virtual,
  title={{Virtual model control: An intuitive approach for bipedal locomotion}},
  author={Pratt, Jerry and others},
  journal={Int. J. Robot. Res},
  volume={20},
  number={2},
  pages={129--143},
  year={2001},
  publisher={SAGE Publications}
}

@article{larby2024optimal,
  title={{Optimal Virtual Model Control for Robotics: Design and Tuning of Passivity-Based Controllers}},
  author={Larby, Daniel and Forni, Fulvio},
  journal={arXiv:2411},
  year={2024}
}

@inproceedings{yang2021reactive,
  title={Reactive human-to-robot handovers of arbitrary objects},
  author={Yang, Wei and others},
  booktitle={2021 IEEE Int. Conf. Robot. Autom. (ICRA)},
  pages={3118--3124},
}

@article{duan2024reactive,
  title={{Reactive Human-to-Robot Dexterous Handovers for Anthropomorphic Hand}},
  author={Duan, Haonan and others},
  journal={IEEE Trans. Robot.},
  year={2024},
  publisher={IEEE}
}

@misc{VMRobotControl,
    author = {Daniel Larby},
    title = {{VMRobotControl.jl}},
    howpublished = {\url{https://github.com/Cambridge-Control-Lab/VMRobotControl}},
    year = {2024}
}

@article{huang2023fed,
  title={Fed-HANet: Federated visual grasping learning for human robot handovers},
  author={Huang, Ching-I and others},
  journal={IEEE Robot. Autom. Lett.},
  volume={8},
  number={6},
  pages={3772--3779},
  year={2023},
  publisher={IEEE}
}

@inproceedings{djeha2022human,
  title={Human-robot handovers using task-space quadratic programming},
  author={Djeha, Mohamed and others},
  booktitle={2022 31st IEEE Inter. Conf. on Robot and Human Inter. Comm. (RO-MAN)},
  pages={518--523},
}

@article{costanzo2021handover,
  title={Handover control for human-robot and robot-robot collaboration},
  author={Costanzo, Marco and De Maria, Giuseppe and Natale, Ciro},
  journal={Front. Robot. AI},
  volume={8},
  pages={672995},
  year={2021},
  publisher={Frontiers Media SA}
}

@article{pang2025stereo,
  title={Stereo Hand-Object Reconstruction for Human-to-Robot Handover},
  author={Pang, Yik Lung and Xompero, Alessio and Oh, Changjae and Cavallaro, Andrea},
  journal={IEEE Robot. Autom. Lett.},
  year={2025},
  publisher={IEEE}
}

\end{document}